%% file: Template.tex
\documentclass{article}
\usepackage{spconf,amsmath,graphicx,hyperref}
\usepackage[table]{xcolor}
\usepackage{amsfonts}

\title{Teacher-Student Diffusion Model for Text-Driven 3D Hand Motion Generation}
\name{
Ching-Lam Cheng \quad
Bin Zhu\sthanks{Corresponding author.} \quad
Shengfeng He
}
\address{
cl.cheng.2024@phdcs.smu.edu.sg \quad
binzhu@smu.edu.sg \quad
shengfenghe@smu.edu.sg \\
Singapore Management University
}

\vspace{-10mm}
\usepackage{fancyhdr}
\begin{document}
%\ninept
%

\maketitle
\input{sec/0_abstract}

% \vspace{-5mm}
\input{sec/1_intro}
\input{sec/2_related_work}
\input{sec/3_methodology}
\input{sec/4_experiment}

\input{sec/5_conclusion}

\clearpage
% \vspace{-5mm}
\section{Acknowledgements}  
\vspace{-3mm}
This work is supported by the Singapore Ministry of Education Academic Research Fund Tier 1 (Proposal ID: 24-SIS-SMU-015).
\vspace{-5mm}

\bibliographystyle{IEEEbib}
\bibliography{strings,main}

\end{document}

%% file: sec/0_abstract.tex
\begin{abstract}
Generating realistic 3D hand motion from natural language is vital for VR, robotics, and human-computer interaction. Existing methods either focus on full-body motion, overlooking detailed hand gestures, or require explicit 3D object meshes, limiting generality. We propose TSHaMo, a model-agnostic teacher–student diffusion framework for text-driven hand motion generation. The student model learns to synthesize motions from text alone, while the teacher leverages auxiliary signals (e.g., MANO parameters) to provide structured guidance during training. A co-training strategy enables the student to benefit from the teacher’s intermediate predictions while remaining text-only at inference. Evaluated using two diffusion backbones on GRAB and H2O, TSHaMo consistently improves motion quality and diversity. Ablations confirm its robustness and flexibility in using diverse auxiliary inputs without requiring 3D objects at test time.

% Hand motion generation is an area with a wide range of applications, from robotics to virtual reality. Many researchers focus on generating full human body motion or subsets of hand motion, such as hand-object interaction, which requires some form of additional input other than a text prompt. This paper focuses on diffusion models, which have demonstrated astonishing results on generation tasks in many domains, as well as motion generation tasks. By pairing two transformer-based diffusion models together, where one is guided by the other, which is conditioned on auxiliary information, the student model can learn from the teacher's path and reach a better optimum. We compare our results with those of a basic diffusion model and baselines and demonstrate significant enhancements in the evaluation metrics. We also analyze the effect of tuning hyperparameters for guidance, demonstrating the guidance model's impact on the original diffusion process.                                        

\end{abstract}

%% file: sec/1_intro.tex
\section{Introduction}
\label{sec:intro}
% \vspace{-4mm}

Human hands are essential for communication, manipulation, and interaction. Generating realistic hand motion from text (\textit{text-to-3D hand motion}) benefits VR~\cite{burdea2003virtual}, prosthetics~\cite{geethanjali2016myoelectric}, and robotic~\cite{billard2019trends}s. Text-driven motion generation spans full-body, hand, and hand–object interaction. Full-body methods~\cite{T2M,MDM,imos} often neglect fine-grained hand dynamics, while hand–object approaches~\cite{text2hoi,diffh2o, tang2025ragg} rely on 3D object meshes, limiting applicability. Yet in many real-world settings (e.g., AR/VR, gesture communication), only text is available, making object-free hand motion generation both valuable and underexplored.

We propose \textbf{TSHaMo}, a model-agnostic \textbf{T}eacher-\textbf{S}tudent diffusion model for text-to-3D \textbf{HA}nd \textbf{MO}tion generation. The student learns to synthesize motion from text alone, ensuring flexible deployment. The teacher, enriched with auxiliary cues (e.g., MANO parameters, joints, contact maps), provides structured supervision. Through co-training, the student benefits from both ground-truth and teacher outputs, improving realism and semantic alignment. TSHaMo is architecture-agnostic and can be integrated into diverse diffusion models.

Unlike hand-object interaction methods~\cite{text2hoi}, our framework requires no object mesh, enabling general-purpose hand motion synthesis. We evaluate on H2O~\cite{h2Odataset} and GRAB~\cite{GRAB}, showing consistent improvements over MDM~\cite{MDM} and StableMoFusion~\cite{huang2024stablemofusion}, with ablations confirming robustness to auxiliary types and hyperparameters.

The contribution of the paper is summarized as follows:
\vspace{-2mm}
\begin{itemize}
    \item A model-agnostic teacher-student diffusion framework for text-to-3D hand motion generation without object meshes. \vspace{-3mm}
    \item A co-training strategy where the teacher provides auxiliary-guided supervision without altering student inference. \vspace{-3mm}
    \item Empirical validation on GRAB and H2O, improving two diffusion baselines. \vspace{-3mm}
    \item Extensive ablations showing flexibility across auxiliary conditions and hyperparameters.
\end{itemize}

\vspace{-5mm}

%% file: sec/2_related_work.tex
\section{Related Work}
\label{sec:related_work}

% \vspace{-3mm}

\subsection{Text-to-Human Motion Generation}
Recent years have seen rapid progress in text-driven human motion generation~\cite{T2M,unimotion,zhong2023attt2m,jiang2024motiongpt}. Early works such as T2M~\cite{T2M} adopt VAE-based frameworks for diverse motion synthesis. UniMotion~\cite{unimotion} enables both global and per-frame text control.  

Diffusion models~\cite{ddpm} have become dominant, with MDM~\cite{MDM}, MotionDiffuse~\cite{motiondiffuse}, and StableMoFusion~\cite{huang2024stablemofusion} achieving state-of-the-art realism and diversity. MDM and MotionDiffuse employ transformer-based architectures~\cite{transformer}, whereas StableMoFusion explores alternatives from Conv1D U-Net to RetNet.  
Although hands are part of full-body motion, these works primarily target global body movements (e.g., walking, running) and lack fine-grained hand gesture modeling.

\vspace{-3mm}

\subsection{Text-to-Hand Motion Generation}
Research on text-driven hand motion is limited, with most focusing on hand-object interaction~\cite{text2hoi,diffh2o}. Early works generate interaction given object meshes~\cite{taheri2021goal}, later extended with text for added control. Text2HOI~\cite{text2hoi} enforces realism via predicted 3D contact maps. While effective, these methods require explicit object inputs and cannot model general hand motions independent of objects.

\vspace{-3mm}

\subsection{Learning from Auxiliary Knowledge}
Leveraging auxiliary signals to improve training has been widely studied. Bellitto et al.~\cite{bellitto2022effectsauxiliaryknowledgecontinual} use auxiliary knowledge to enhance continual learning, while Guo et al.~\cite{T2M} combine posterior and prior networks. Such approaches show that auxiliary information can guide models toward better optima without changing their inference conditions.

\vspace{-2mm}

%% file: sec/3_methodology.tex
\newcommand{\boldx}{\textbf{x}}
\newcommand{\loss}{\mathcal{L}}

\begin{figure*}[t]
  \centering
  
   \includegraphics[width=1.0\linewidth]{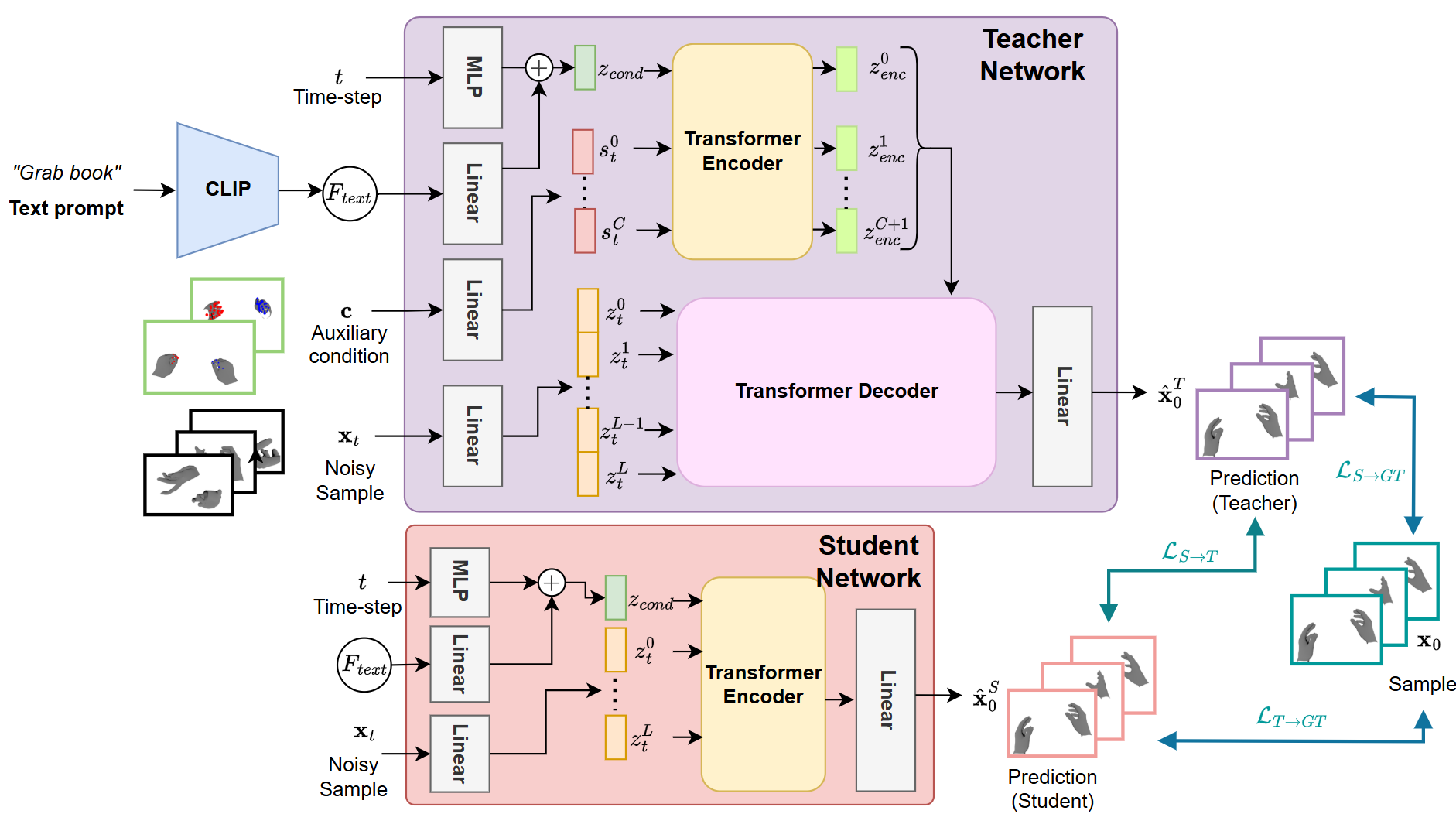}
   \vspace{-8mm}
    \caption{Training procedure of our method. The student takes a noisy sample and text embedding to predict the denoised output, while the teacher also uses auxiliary conditions (e.g., 3D joints, MANO parameters, contact maps). Three losses are applied between model predictions and ground truth. The example here uses the MDM backbone.}

   \label{fig:training_overview}
   \vspace{-4mm}
\end{figure*}

\section{Methodology}
\vspace{-2mm}
\label{sec:methodology}

\subsection{Preliminaries}
\vspace{-2mm}
\subsubsection{Problem Formulation.}
\vspace{-2mm}
Given a text prompt $\textbf{T}$, the goal is to generate coherent 3D hand motion sequences $\tilde\boldx_0 = \{\tilde\boldx_0^i\}_{i=0}^L$ aligned with the prompt. Each frame represents a static hand pose parameterized by MANO~\cite{mano}, including translation $t \in \mathbb{R}^3$, pose $\theta \in \mathbb{R}^{48}$, and shape $\beta \in \mathbb{R}^{10}$. The MANO layer produces hand mesh vertices $\textbf{V} \in \mathbb{R}^{778\times 3}$ and joints $\textbf{J} \in \mathbb{R}^{21\times 3}$. The model also predicts joint-level contact maps, where values indicate inverse hand–object distances. A complete motion sequence is
\vspace{-2mm}
\[
\boldx = \{\textbf{1}_{lhand}^i, x_{lhand}^i, x_{lcont}^i, \textbf{1}_{rhand}^i, x_{rhand}^i, x_{rcont}^i\}_{i=0}^{N}, \vspace{-2mm}
\] 
with $\textbf{1}_{lhand}^i, \textbf{1}_{rhand}^i$ denoting hand presence, $x_{lhand}^i, x_{rhand}^i$ the MANO parameters, and $x_{lcont}^i, x_{rcont}^i \in \mathbb{R}_{[0,1]}^{21}$ the contact maps.
\vspace{-2mm}
\subsubsection{Diffusion Models.}
\vspace{-2mm}
The diffusion process~\cite{ddpm} gradually corrupts $\boldx_0$ with Gaussian noise: \vspace{-2mm}
\[
q(x_t \mid x_{t-1}) = \mathcal{N}(x_t; \sqrt{1-\beta_t}\,x_{t-1}, \beta_t \mathbf{I}), \vspace{-2mm}
\]
with $\beta_t$ controlling noise scale. The reverse process is modeled as
\vspace{-2mm}
\[
p_\theta(x_{t-1}\mid x_t) = \mathcal{N}(x_{t-1}; \mu_\theta(x_t,t), \Sigma_\theta(x_t,t)), \vspace{-2mm}
\]
starting from $x_T \sim \mathcal{N}(\mathbf{0}, \mathbf{I})$. Following Tevet et al.~\cite{MDM}, we predict $\boldx_0$ directly instead of noise.
\vspace{-2mm}
\subsection{Teacher-Student Diffusion Model}
\vspace{-2mm}
\label{sec:teacher_student}
Our framework (Figure~\ref{fig:training_overview}) uses a teacher network $f^T$ and a student $f^S$. The student generates motion from text only, while the teacher leverages both text and auxiliary cues (e.g., MANO, joints, contact maps).

\noindent \textbf{Teacher.} The teacher network $f^T$ takes text prompts, timestep embeddings, and auxiliary features ${s_t^0 \dots s_t^C}$ as input, and outputs the denoised motion $\hat{\boldx}_0^T$. It can be built on different diffusion backbones; for example, with MDM~\cite{MDM}, we use a transformer encoder–decoder where CLIP-encoded text and timestep embeddings form $z_{cond}$, concatenated with auxiliary features and processed by the encoder, while the decoder denoises $\boldx_t$.

\noindent \textbf{Student.} The student network $f^S$ maps text prompts, timestep embeddings, and noisy motion $\boldx_t$ to $\hat{\boldx}_0^S$. With MDM as backbone, $z_{cond}$ is formed from CLIP features and timestep embeddings, concatenated with $\boldx_t$, and passed through a transformer encoder to produce the output.

\noindent \textbf{Model-Agnostic Design.} TSHaMo is compatible with different diffusion backbones. For MDM, we extend to an encoder–decoder for auxiliary inputs. For StableMoFusion~\cite{huang2024stablemofusion}, we add input channels for auxiliary signals without altering the student. In both, the student remains unchanged, demonstrating generality.

\vspace{-3mm}
\subsection{Co-training}
\vspace{-2mm}
\label{sec:cotraining}
We jointly optimize student and teacher with complementary losses:
\vspace{-2mm}
\[
\loss_S = \loss_{S\rightarrow GT} + \lambda \loss_{S\rightarrow T}, \quad 
\loss_T = \loss_{T \rightarrow GT}, \vspace{-2mm}
\]
where $\lambda$ balances teacher guidance. Denoising losses are
\vspace{-2mm}
\[
\loss_{. \rightarrow GT} = \mathbb{E}||\boldx_0 - f^.(\boldx_t,t,c_.)||^2_2, . \in \{S,T\} \vspace{-2mm}
\]
and teacher supervision is
\vspace{-2mm}
\[
\loss_{S \rightarrow T} = \mathbb{E}||f^S(\boldx_t,t,c_S) - f^T(\boldx_t,t,c_T)||^2_2. \vspace{-2mm}
\]

To avoid premature convergence, $f^T$ is updated less frequently using a hyperparameter $t_{cycle}$. During training, a count-down timer is used, where each time the teacher model update, the timer is reset to $t_{counter} = \lfloor t_{cycle}\rfloor + \textbf{1}_{u \leq t_{cycle}- \lfloor t_{cycle}\rfloor}$, where $u$ is sampled from $\mathcal{U}(0,1)$, such that $\mathbb{E}[t_{counter}] = t_{cycle}$ steps. Both models adopt classifier-free guidance~\cite{cfg}, enabling trade-offs between accuracy and diversity by scaling:
\vspace{-2mm}
\[
f_{sample}^{\cdot}(\boldx_t, t, c) = f^{\cdot}(\boldx_t, t, \emptyset) + \sigma\,(f^{\cdot}(\boldx_t, t, c) - f^{\cdot}(\boldx_t, t, \emptyset)). \vspace{-2mm}
\]

During inference, only the student is used with text prompts. Sampling begins with $\boldx_T \sim \mathcal{N}(\mathbf{0},\mathbf{I})$ and iteratively denoises until $\tilde\boldx_0$ is obtained.

\vspace{-2mm}

%% file: sec/4_experiment.tex
\newcommand{\confint}[2]{$#1^{\pm #2}$}
\newcommand{\confintb}[2]{$\textbf{#1}^{\pm \textbf{#2}}$}

\begin{table*}
    \centering
    \caption{3D hand motion generation performance comparison on the GRAB dataset. The best results are highlighted in bold.}
    \begin{tabular}{lcccccc} \hline
        Method& Acc(Top 1)$\uparrow$ &Acc(Top 2)$\uparrow$&Acc(Top 3)$\uparrow$& KID  ($\times 5000$)$\downarrow$& Diversity\\ \hline 
        \rowcolor{gray!5}
        GT&\confint{0.968}{0.004}& \confint{0.976}{0.006}& \confint{0.983}{0.007} & - &  \confint{0.670}{0.001} \\ \hline

        \rowcolor{green!5}
        T2M~\cite{T2M} &\confint{0.121}{0.030}&\confint{0.210}{0.033}&\confint{0.273}{0.029} &\confint{1.599}{0.062}&\confint{0.444}{0.014}\\
        
        \rowcolor{orange!5}
        MDM~\cite{MDM}&\confint{0.500}{0.016}& \confint{0.572}{0.031}&\confint{0.645}{0.029}& \confint{0.166}{0.013}&\confint{0.594}{0.009}\\
        
        \rowcolor{blue!5} StableMoFusion~\cite{huang2024stablemofusion}&\confint{0.710}{0.022}&\confint{0.887}{0.018}&\confintb{0.935}{0.021} &\confint{0.130}{0.006}&\confint{0.753}{0.012}\\ \hline
        \rowcolor{orange!5}
        \textbf{TSHaMo (MDM)} &\confint{0.806}{0.017}&\confint{0.846}{0.014}& \confint{0.878}{0.013}  &\confint{0.122}{0.006}& \confint{0.658}{0.010} \\
        \rowcolor{blue!5}
        \textbf{TSHaMo (StableMoFusion)}&\confintb{0.834}{0.032}&\confintb{0.887}{0.024}&\confint{0.903}{0.012} &\confintb{0.114}{0.004}&\confint{0.740}{0.011}\\
        \hline
    \end{tabular}
    \label{tab:overall_result_GRAB}
    \vspace{-5mm}
\end{table*}

\begin{table*}
    \centering
    \caption{Comparison of 3D hand motion generation performance on the H2O dataset. Best results are shown in bold.}
    \begin{tabular}{lccccc} \hline
        Method & Acc(Top 1) & Acc(Top 2) & Acc(Top 3)$\uparrow$ & KID ($\times 5000$) $\downarrow$ & Diversity \\ \hline 
        \rowcolor{gray!5}
        GT & \confint{0.750}{0.000} & \confint{0.922}{0.000} & \confint{0.948}{0.000} & - & \confint{0.470}{0.000} \\ \hline
        \rowcolor{green!5}
        T2M~\cite{T2M} & \confint{0.215}{0.033} & \confint{0.396}{0.030} & \confint{0.561}{0.028} & \confint{1.974}{0.046} & \confint{0.178}{0.009} \\
        \rowcolor{orange!5}
        MDM~\cite{MDM} & \confint{0.207}{0.011} & \confint{0.346}{0.039} & \confint{0.527}{0.026} & \confint{1.243}{0.023} & \confint{0.434}{0.008} \\
        \rowcolor{blue!5}
        StableMoFusion~\cite{huang2024stablemofusion} & \confint{0.207}{0.024} & \confint{0.517}{0.018} & \confint{0.594}{0.020} & \confint{2.341}{0.136} & \confint{0.722}{0.008} \\ \hline
        \rowcolor{orange!5}
        \textbf{TSHaMo(MDM)} & \confint{0.206}{0.025} & \confint{0.483}{0.036} & \confint{0.610}{0.015} & \confintb{1.093}{0.035} & \confint{0.403}{0.010} \\
        \rowcolor{blue!5}
        \textbf{TSHaMo(StableMoFusion)} & \confintb{0.258}{0.028} & \confintb{0.577}{0.010} & \confintb{0.655}{0.016} & \confint{1.243}{0.041} & \confint{0.841}{0.013} \\
        \hline
    \end{tabular}
    \label{tab:overall_result}
    \vspace{-4mm}
\end{table*}

\begin{table*}
    \centering
    \caption{Ablation study for auxiliary condition types.}
    \resizebox{\textwidth}{!}{
    \begin{tabular}{lccc||ccc} \hline
        &&Student &&&Teacher\\
        Cond. type&  Accuracy (Top 3) $\uparrow$&   KID  ($\times 5000$)$\downarrow$ & Diversity &  Accuracy (Top 3) $\uparrow$&   KID  ($\times 5000$)$\downarrow$& Diversity\\ \hline 
        No cond (MDM) &\confint{0.527}{0.026}&\confint{1.243}{0.023}& \confint{0.455}{0.009} & - & - & - \\ \hline
        MANO params & \confint{0.591}{0.022}  & \confint{1.102}{0.022} & \confint{0.399}{0.008} & \confint{0.883}{0.010} & \confint{0.053}{0.003} & \confint{0.470}{0.013} \\
        3D joints & \confint{0.596}{0.030} & \confint{1.158}{0.022} & \confint{0.434}{0.008} & \confint{0.704}{0.017} & \confint{0.143}{0.039} & \confint{0.428}{0.010} \\
        2D joints & \confint{0.561}{0.020} & \confint{1.192}{0.027} &\confint{0.411}{0.009} & \confint{0.638}{0.015} & \confint{0.266}{0.047} & \confint{0.319}{0.009} \\
        Hand masks & \confint{0.562}{0.017} & \confint{1.145}{0.028} & \confint{0.435}{0.009} & \confint{0.599}{0.027} & \confint{0.106}{0.022} & \confint{0.472}{0.015} \\
        MANO + Hand contact & \confintb{0.610}{0.015} &\confintb{1.093}{0.035} & \confint{0.403}{0.010} & \confintb{0.896}{0.013} & \confintb{0.041}{0.001} & \confint{0.473}{0.010} \\
        \hline
    \end{tabular}}
    \label{tab:condtype_result}
    \vspace{-5mm}
    % \vspace{-1mm}
\end{table*}

\vspace{-1.5mm}
\section{Experiments}
\label{sec:experiment}
\vspace{-2mm}
\subsection{Experiment Settings}
\vspace{-2mm}
\noindent \textbf{Datasets.} We evaluate on two hand–object interaction datasets, H2O~\cite{h2Odataset} and GRAB~\cite{GRAB}. Only hand motion is used, with action labels as text prompts. Object information is discarded during inference to ensure general-purpose generation.  

\noindent \textbf{Evaluation metrics.} We report Top-k accuracy, Kernel Inception Distance (KID), and diversity. Top-k accuracy uses a pretrained action classifier to check if the correct class appears among the top-$k$ predictions (Top-1–3 overall, Top-3 in ablations~\cite{text2hoi}). KID measures distributional realism via squared Maximum Mean Discrepancy with a polynomial kernel, chosen over FID for its unbiasedness and reliability with small samples. Diversity captures variability across generated motions. All metrics are averaged over 20 runs with 95\% confidence intervals.

\noindent \textbf{Baselines.} We compare our method against two state-of-the-art text-to-motion generation methods
T2M~\cite{T2M}, MDM~\cite{MDM} and StableMoFusion~\cite{huang2024stablemofusion}. T2M uses a temporal variational autoencoder (VAE) to synthesize a diverse set of human motions, while MDM and StableMoFusion are classifier-free diffusion-based generative models. MDM uses a transformer-encoder-based architecture, and StableMoFusion uses a U-Net-based architecture. Since all methods are designed for human motion, we retrained them with the hand motion data, enabling them to generate hand motion for fair comparison.

\noindent \textbf{Auxiliary conditioning.} We have chosen MANO parameters and hand contact map as the auxiliary condition for the teacher model to guide the generation process. MANO parameters represent hand pose with global translation, pose, and shape parameters, while the hand contact map indicates the distances between hand joints and the interacting object

\begin{figure}
  \centering
   \includegraphics[width=\linewidth]{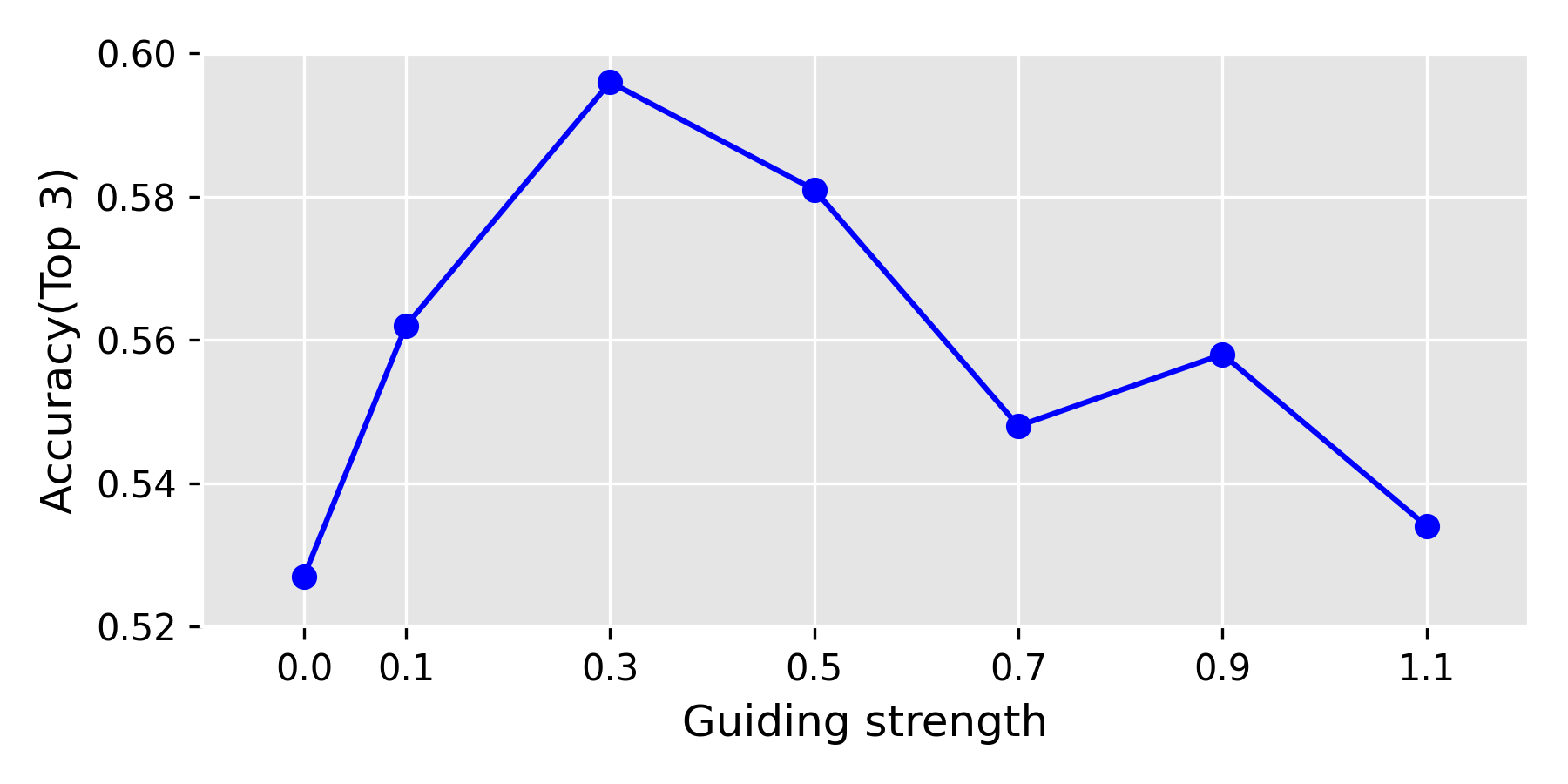}
   \vspace{-10mm}
   \caption{Ablation study of guidance strengths $\lambda$ using 3D hand joints as condition.}
   \label{fig:guidstr}
   \vspace{-5mm}
\end{figure}

% Added model architecture of teacher model for MDM and StableMoFusion
\noindent \textbf{Implementation details.}  
We use $T_{max}=1000$ diffusion steps and train both base and guidance models for 1000 epochs with learning rate $\alpha=10^{-5}$ under a cosine annealing schedule. In inference, cfg scale $\sigma=10.0$ provides the best trade-off between accuracy and fidelity. We set the teacher guidance scale $\lambda$ to 0.3, which yields the best result as shown in Figure~\ref{fig:guidstr}. For auxiliary conditioning, we uniformly sample 5 frames for each sequence, at the 0th, 25th, 50th, 75th, and 100th percentiles. The sampled frames are projected and concatenated with text embedding.
\vspace{-5mm}
\subsection{Performance Comparison}
\vspace{-2mm}
As shown in Tables~\ref{tab:overall_result_GRAB} and \ref{tab:overall_result}, we evaluate TSHaMo on two diffusion-based baselines, MDM~\cite{MDM} and StableMoFusion~\cite{huang2024stablemofusion}, using the GRAB and H2O datasets respectively. On GRAB, TSHaMo improves MDM in top-1 to top-3 accuracy and StableMoFusion in top-1 to top-2. On H2O, it enhances MDM in top-2 and top-3, and StableMoFusion across top-1 to top-3. TSHaMo also reduces KID, showing closer alignment to real motion distributions, while maintaining high diversity. These results demonstrate that TSHaMo generates semantically accurate yet diverse hand motions, validating its effectiveness for text-to-3D hand motion generation.

\begin{figure}
  \centering
   \includegraphics[width=0.9\linewidth]{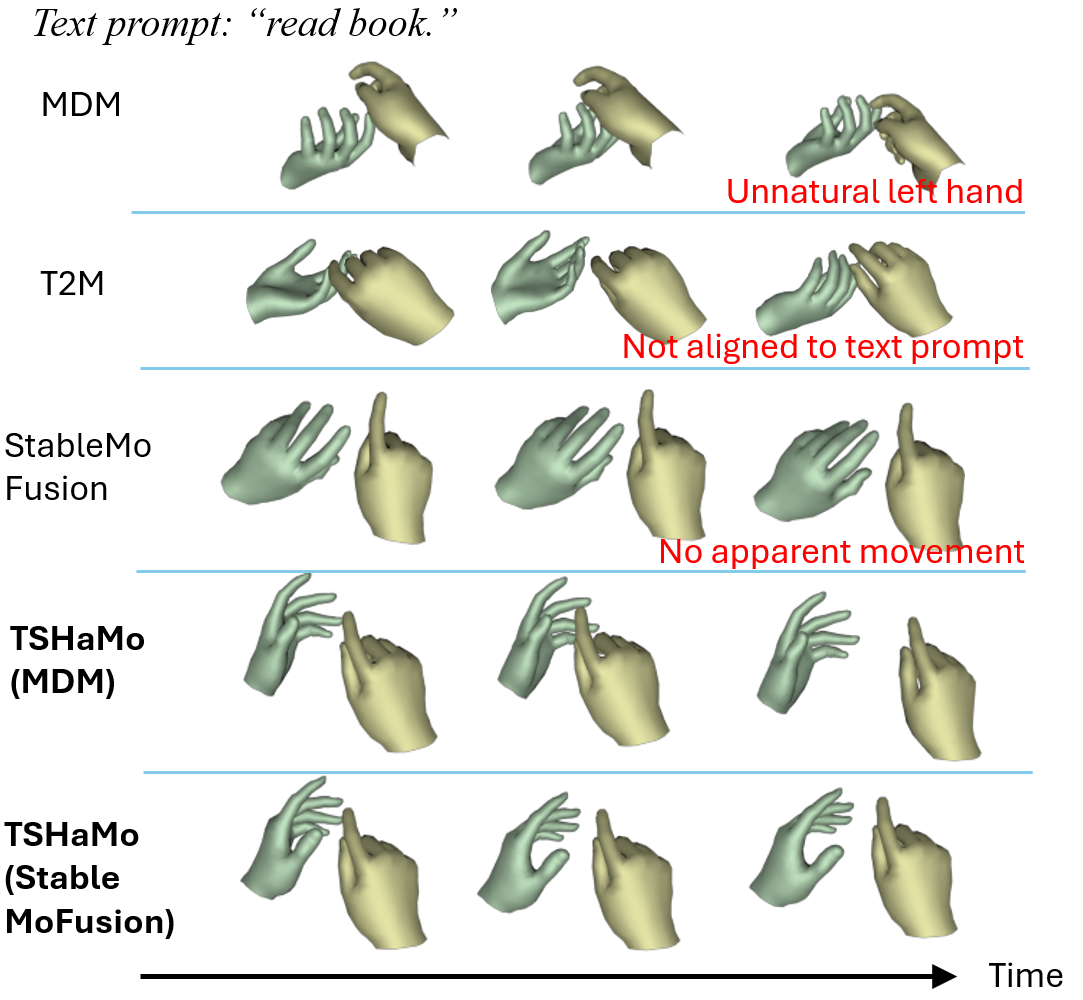}
   \vspace{-5mm}
   \caption{Qualitative example comparison for hand motion generation from textual prompt.}
   \label{fig:qualitative_res}
   \vspace{-5mm}
\end{figure}

% \begin{figure}
%   \centering
%    \includegraphics[width=1.0\linewidth]{figure/qual_res_wacv/qr_p2.png}
%    \caption{Hand motions generated by TSHaMo for various text prompts.}
%    \label{fig:qualitative_res_2}
%    \vspace{-7mm}
% \end{figure}
\vspace{-5mm}

\subsection{Ablation Study}
\vspace{-2mm}
We analyze auxiliary conditions and guidance strength $\lambda$ on H2O using MDM as the base model.  

\noindent \textbf{Auxiliary conditions.}  
Table~\ref{tab:condtype_result} shows that combining MANO parameters with contact maps achieves the best accuracy, while MANO + 3D joints also perform strongly. Even partial cues (e.g., 2D joints, image masks) provide useful guidance, highlighting the framework’s flexibility.  

\noindent \textbf{Guidance strength.}  
Figure~\ref{fig:guidstr} shows accuracy rising with $\lambda$ up to 0.3, then declining, indicating that moderate guidance helps, but excessive guidance over-constrains the model.

% \noindent \textbf{Number of frames.}
% Table~\ref{tab:condlen_result} exhibits the result for different numbers of frames $l_{cond}$ used in auxiliary conditions. As the number of frames for the condition increases, the performance of the teacher model improves steadily as guiding strength increases. 
% % This coincides with our expectation, as the guidance model is receiving more details as the frame length increases.
% However, such a trend does not reflect on the student model, as varied results are shown on different $l_{cond}$, with $l_{cond}=5$ showing the best performance,  suggesting a trade-off between information richness and training complexity. All student model, except for those with $l_{cond}=1$ and $l_{cond}=2$, shows significant improvement compared to the MDM baseline, where there is no guidance from the teacher model. The results are not surprising as shorter lengths (e.g., 1 or 2) provide insufficient temporal coverage for guiding motion generation. 

% \noindent \textbf{Update cycle.}
% Table~\ref{tab:bupdate_result} shows the influence of the update cycle time when using the MANO parameter as auxiliary condition. The setting $t_{cycle}=1.7$  achieves the best student performance, suggesting that decoupling teacher and student learning speeds prevents premature convergence and maintains meaningful supervision throughout training.

\vspace{-5mm}
\subsection{Qualitative Result}
\vspace{-2mm}
We present qualitative comparisons between our TSHaMo and the three baseline models in Figure~\ref{fig:qualitative_res}. The visualizations clearly demonstrate the superiority of our approach in generating semantically aligned and physically plausible hand motions in response to text prompts, validating the effectiveness of our teacher-guided training strategy in improving the precision and naturalness of hand motion generation.

\vspace{-2mm}

%% file: sec/5_conclusion.tex
\section{Conclusion}
\label{sec:conclusion}
\vspace{-2mm}

In this paper, we introduced TSHaMo, a teacher–student diffusion framework for text-driven 3D hand motion generation. By guiding a text-only student with a teacher enriched by auxiliary signals, TSHaMo enables expressive, object-agnostic gesture synthesis and integrates seamlessly with diffusion backbones such as MDM and StableMoFusion. Experiments on GRAB and H2O demonstrate consistent gains in accuracy, realism, and diversity, while ablations confirm its flexibility across auxiliary inputs and training setups. Future work will explore richer language inputs, including multi-sentence and temporally grounded instructions, for more coherent interactive motion generation.